\documentclass[runningheads]{llncs}
\usepackage{graphicx}
\usepackage{amsmath,amssymb} 
\usepackage{ruler}
\usepackage{color}
\usepackage[width=122mm,left=12mm,paperwidth=146mm,height=193mm,top=12mm,paperheight=217mm]{geometry}
\usepackage{hyperref}
\usepackage{bbm}
\usepackage{paralist}
\usepackage{color,soul}
\usepackage{makecell}
\usepackage{multirow, booktabs}
\usepackage{tablefootnote}
\usepackage{subfigure}
\usepackage[font={small}]{caption}

\begin{document}

\providecommand{\ie}{i.e.\ }
\providecommand{\eg}{e.g.\ }
\providecommand{\aka}{a.k.a.\ }
\providecommand{\hessam}[1]{{\protect\color{red}{\bf [Hessam: #1]}}}
\providecommand{\mch}[1]{{\protect\color{red}{\bf [Max: #1]}}}
\providecommand{\mohammad}[1]{{\protect\color{red}{\bf [Mohammad: #1]}}}
\providecommand{\change}[1]{\hl{#1}}
\providecommand{\ali}[1]{{\protect\color{red}{\bf [Ali: #1]}}}
\providecommand{\up}[1]{{\textsuperscript{#1}}}
\providecommand{\sub}[1]{{\textsubscript{#1}}}
\pagestyle{headings}
\mainmatter

\title{\vspace{-2mm}Label Refinery: Improving ImageNet Classification through Label Progression\vspace{-5mm}} 

\titlerunning{Label Refinery}

\authorrunning{Hessam Bagherinezhad, Maxwell Horton, Mohammad Rastegari, Ali Farhadi}

\author{Hessam Bagherinezhad\up{1,2}, Maxwell Horton\up{1,2}, Mohammad Rastegari\up{1,3}, \\Ali Farhadi\up{1,2,3} \\
\texttt{\{hessam,max,mohammad,ali\}@xnor.ai}}
\institute{\up{1}XNOR AI, \up{2}University of Washington, \up{3}Allen AI}

\maketitle

\begin{figure}[h!]
\vspace{-7mm}
\centering
\includegraphics[width=\textwidth]{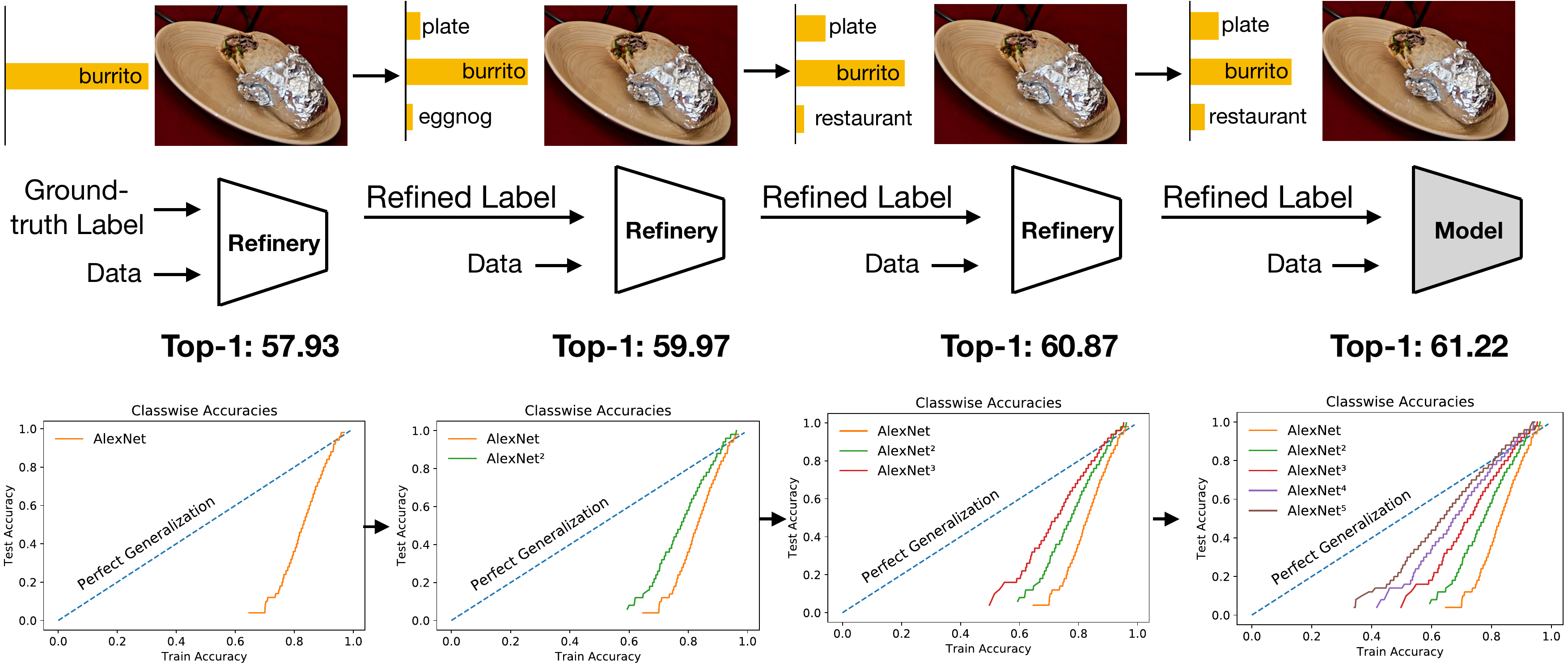}
\caption{\footnotesize Current labeling principles impose challenges for machine learning models. We introduce the \textit{Label Refinery}, an iterative procedure to update ground truth labels using a visual model trained on the entire dataset. The Label Refinery produces soft, multi-category, dynamically-generated labels consistent with the visual signal. The training image shown is labelled with the single category ``burrito''. After a few iterations of label refining, the labels from which the final model is trained are informative, unambiguous, and smooth. This results in major improvements in the model accuracy during successive stages of refinement as well as improved model generalization. These plots show that as models proceed through successive stages of refinement, the gaps between train and test results and approach ideal generalization.}
\label{fig:teaser}
\end{figure}

\begin{abstract}
\vspace{-1cm}
Among the three main components (data, labels, and models) of any supervised learning system, data and models have been the main subjects of active research. However, studying labels and their properties has received very little attention. Current principles and paradigms of labeling impose several challenges to machine learning algorithms. Labels are often incomplete, ambiguous, and redundant. In this paper we study the effects of various properties of labels and introduce the \textit{Label Refinery}: an iterative procedure that updates the ground truth labels after examining the entire dataset. We show significant gain using refined labels across a wide range of models. Using a Label Refinery improves the state-of-the-art top-1 accuracy of (1) AlexNet from $59.3$ to $67.2$, (2) MobileNet\up{1} from $70.6$ to $73.39$, (3) MobileNet\up{0.25} from $50.6$ to $55.59$, (4) VGG19 from $72.7$ to $75.46$, and (5) Darknet19 from $72.9$ to $74.47$.

\end{abstract}

\section{Introduction}
\par
There are three main components in the typical pipeline of supervised learning systems: the \textit{data}, the \textit{model}, and the \textit{labels}. Sources of data have expanded drastically in past several years. We have observed the impact of large-scale datasets for several visual tasks. 
A variety of data augmentation methods~\cite{wang2017effectiveness,wong2016understanding,xu2016improved,shrivastava2017gan} have effectively expanded these datasets and improved the performance of learning systems. Models have also been extensively studied in the literature. Recognition systems have shown improvements by 
increasing the depth of the architectures \cite{he2016deep,simonyan2014very}, introducing new activation and normalization layers \cite{ioffe2015batch,krizhevsky2012imagenet}, and developing optimization techniques and loss functions~\cite{kingma2014adam,xie2016disturblabel}. In contrast to the improvements in data and models, little effort has focused on improving labels. 

\par
\begin{figure}[t]
\centering
\subfigure[]{
  \includegraphics[width=0.3\textwidth]{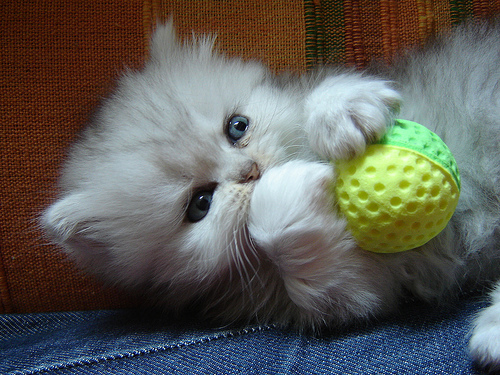}
  \label{fig:persian-cat}
}
\hspace{0.5cm}
\subfigure[]{
  \includegraphics[width=0.2\textwidth]{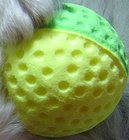}
  \label{fig:persian-cat-cropped}
}
\vspace{-0.4cm}
\caption{\small Figure~\ref{fig:persian-cat} shows a sample image from the ``persian cat" category of ImageNet's training set. The standard technique to train modern state-of-the-art architectures is to crop patches as small as $8\%$ area of the original image, and label them with the original image's label. This will often result in inaccurate labels for the augmented data. Figure~\ref{fig:persian-cat-cropped} shows a sample crop of the original image where the ``persian cat" is no longer in the crop. A trained ResNet-50 labels Figure~\ref{fig:persian-cat} by ``persian cat", and labels Figure~\ref{fig:persian-cat-cropped} by ``golf ball". We claim that using a model to generate labels for the patches results in more accurate labels and therefore more accurate models.}
\label{fig:bad-crop-labels}
\vspace{-0.4cm}
\end{figure}

Current labeling principles and practices impose specific challenges on our learning algorithms. 1) Incompleteness: A natural image of a particular category will contain other object categories as well. For example, Figure~\ref{fig:persian-cat} shows an example from ImageNet that is labeled ``cat" but the image contains a ``ball" as well. This problem is rooted in the nature of how researchers define and collect labels, and is not unique to a specific dataset. 2) Taxonomy Dependency: Categories that are far from each other in the taxonomy structure can be very similar visually. 3) Inconsistency: To prevent overfitting, various loss functions and regularization techniques have been introduced into the training process. Data augmentation~\cite{wang2017effectiveness} is one of the most effective methods employed to prevent neural networks from memorizing the training data. Most modern state-of-the-art architectures for image classification are trained with crop-level data augmentation, in which crops of the image used for training can be as small as $8\%$ of the area of the original image~\cite{he2016deep}. For many categories, such small crops will frequently result in patches in which the object of interest is no longer visible (Figure~\ref{fig:bad-crop-labels}), resulting in an inconsistency with the original label.

\par
To address the aforementioned shortcomings, we argue that several characteristics should apply to ideal labels. Labels should be \textit{soft} to provide more coverage for co-occurring and visually-related objects. Traditional one-hot vector labels introduce challenges in the modeling stage. Labels should be \textit{informative} of the specific image, meaning that they should not be identical for all the images in a given class. For example, an image of a ``dog" that has similar appearance to a ``cat" should have a different label than an image of a ``dog" that has similar appearance to a ``fox". This also suggest that labels should be defined at the instance-level rather than the category-level. Determining the best label for each instance may require observing the entire data to establish intra- and inter-category relations, suggesting that labels should be \textit{collective} across the whole dataset. Labels should also be consistent with the image content when crops are taken. Therefore, labels should be \textit{dynamic} in the sense that the label for a crop should depend on the content of the crop.

\par
In this paper we introduce \textit{Label Refinery}, a solution that uses a neural network model and the data to modify crop labels during training. Refining the labels while training enables us to generate soft, informative, collective, and dynamic labels. Figure~\ref{fig:teaser} depicts an example of a label refinery. As models go through the stages of the refinery labels are updated based on the previous models. This results in major improvements in the accuracy and generalization. The output of the lable refinery is a set of labels from which one can learn a model. The model trained from the produced lables are much more accurate and more robust to overfiting. 

Our experiments show that Label Refining consistently improves the accuracy of object classification networks by a large margin across a variety of popular network architectures. Our improvements in Top-1 accuracy on the ImageNet validation set include: AlexNet from $59.3\%$ to $67.2\%$, VGG19 from $72.7\%$ to $75.46\%$, ResNet18 from $69.57\%$ to $72.52\%$, ResNet50 from $75.7\%$ to $76.5\%$, DarkNet19 from $72.9\%$ to $74.47\%$, MobileNet\sub{0.25} from $50.65\%$ to $55.59\%$, and MobileNet\sub{1} from $70.6\%$ to $73.39\%$. Collective and dynamic labels enable standard models to generalize better, resulting in significant improvements in image classification. Figure~\ref{fig:teaser} Plots the train versus test accuracies as models go through the label refinery procedure. The gap between train and test accuracies is getting smaller and closer to an ideal generalization. 

\par
We further demonstrate that a trained model can serve as a Label Refinery for another model of the same architecture. For example, we iterate through several successions of training a new AlexNet model by using the previously trained AlexNet model as a Label Refiner. Our results show major improvements (from $59.3\%$ to $61.2\%$) on using AlexNet to refine lables for another AlexNet. Note that the final AlexNet has not seen the actual groundtruth labels in the past few stages. The final AlexNet models demonstrate greatly reduced overfitting compared to the original models (Figure~\ref{fig:overfitting_alex} and Figure~\ref{fig:trainval-accuracies-alexnet}). We also experiment with using a model of one architecture as a Label Refiner for a model of another architecture. Further, we have also shown that adversarialy modifying image examples improves the accuracy when using label refinery.
\par
Our contributions include: (1) introducing the Label Refinery for crop-level label augmentation, (2) improving state-of-the-art accuracy on ImageNet for a variety of existing architectures, (3) demonstrating the ability of a network to improve accuracy by training from labels generated by another network of the same architecture, and (4) generating adversarial examples to improve the performance of the Label Refinery method.


\begin{figure}[t]
\centering
\subfigure[]{
  \includegraphics[height=0.15\textwidth]{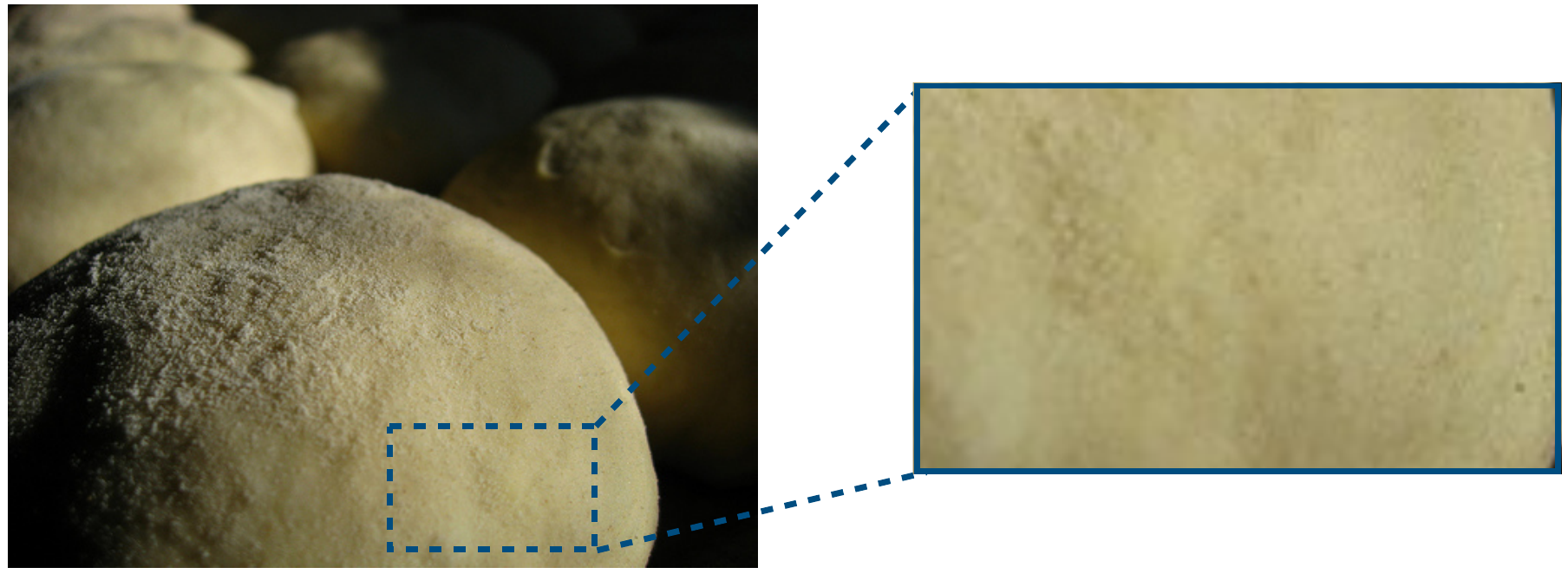}
  \label{fig:dough}
}
\subfigure[]{
  \includegraphics[height=0.15\textwidth]{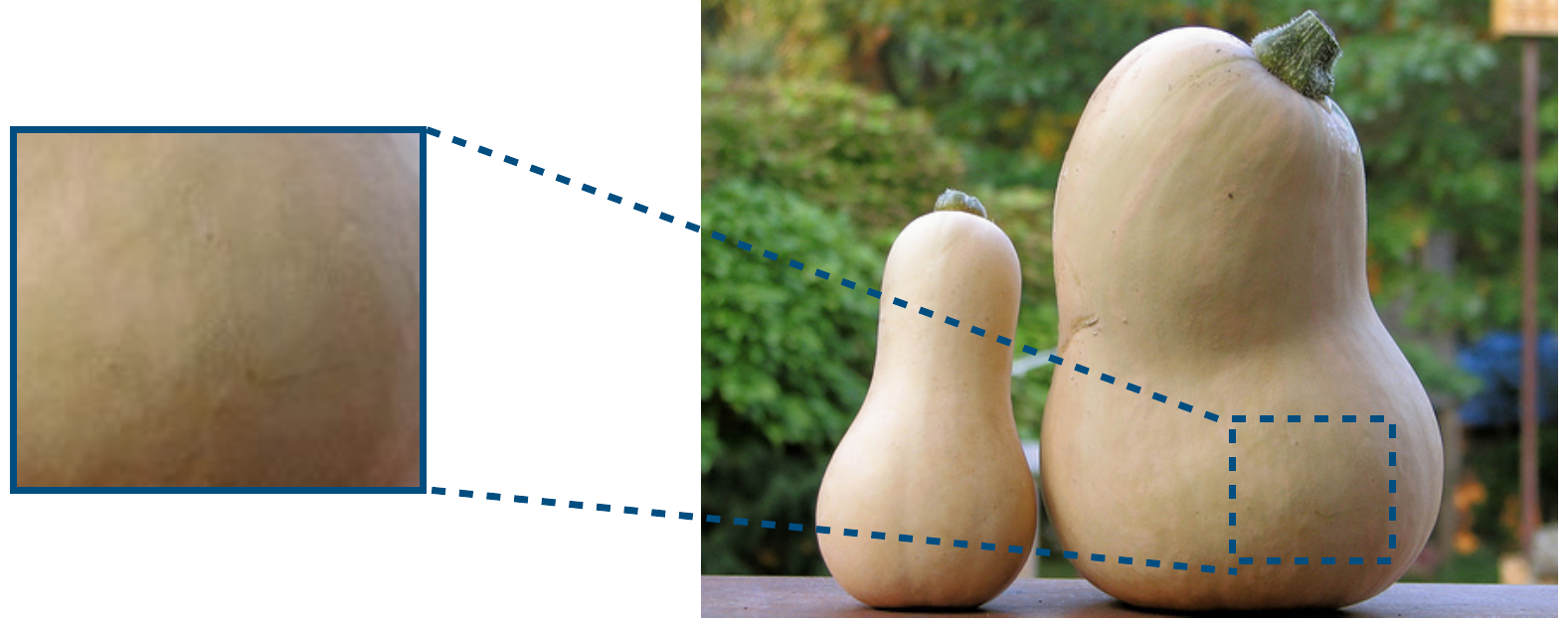}
  \label{fig:butternut-squash}
}
\caption{\small Sample training examples from ``dough" and ``butternut squash" categories of ImageNet. While the two sample images are visually distinctive, their random crops are quiet similar. A trained ResNet-50 labels both cropped patches softly over categories of ``dough", ``butternut squash", ``burrito", ``french loaf", and ``spaghetti squash". We claim that labelling the crops softly by a trained model makes the training of the same model more stable, and therefore results in more accurate models.}
\label{fig:crop-smoothing}
\vspace{-0.4cm}
\end{figure}

\section{Related Work}
\textbf{Label Smoothing and Regularization:} Softening labels has been used to improve generalization. \cite{szegedy2016rethinking} uniformly redistributes $10\%$ of the weight from the ground-truth label to other classes to help regularize during training. DisturbLabel~\cite{xie2016disturblabel} replaces some of the labels in a training batch with random labels. This helps regularize training by preventing overfitting to ground-truth labels. \cite{reed2014training} augments noisy labels using other models to improve label consistency. \cite{miyato2015distributional} introduces a notion of local distributional smoothness in model outputs based on the smoothness of the model's outputs when inputs are perturbed. The smoothness criterion is enforced with the purpose of regularizing models. The work of \cite{pereyra2017regularizing} explores penalizing networks by regularizing the entropy of the model outputs. Unlike our method, these approaches can not address the inconsistency of the labels. 

\textbf{Incorporating Taxonomy:} Several methods have explored using taxonomy to improve label and model quality. \cite{li2017learning} uses cross-category relationships from knowledge graphs to mitigate the issues caused by noisy labels. \cite{wu2017hierarchical} designs a hierarchical loss to reduce the penalty for predictions that are close in taxonomy to the ground-truth. \cite{wu2015ml} investigates learning multi-label classification with missing labels. They incorporate instance-level information as well as semantic hierarchies in their solution. Incorporating taxonomic information directly into the model's architecture is explored in~\cite{cai2007exploiting}. \cite{mcauleyoptimization} uses the output of existing binary classifiers to address the problem of training models on single-label examples that contain multiple training categories. These methods fail to address the incompleteness of the labels. Instead of directly using taxonomy, our model collectively infer the visual relations between categories to impose these knowledge into the training while capturing a complete description of the image.     




\textbf{Data Augmentation:} To preserve generalization, several data augmentations such as cropping, rotating, and flipping input images have been applied in training models~\cite{krizhevsky2012imagenet,simonyan2014very,he2016deep,szegedy2017inception}. \cite{wong2016understanding} proposes data warping and synthetic over-sampling to generate additional training data. \cite{wang2017effectiveness} and \cite{shrivastava2017gan} explore using GANs to generate training examples. Most of such augmentation techniques further confuse the model with inconsistent labels. For example, a random crop of an image might not contain the main object the that image We propose augmenting the labels alongside with the data by refining them during training when augmenting the data.

\textbf{Teacher-Student Training:} Using another network or an ensemble of multiple networks as a teacher model to train a student model has been explored in \cite{caruana2006compression,ba2014deep,li2014learning,shenteacher,hinton2015distilling,romero2014fitnets}. \cite{ba2014deep} explores training a shallow student network from a deeper teacher network. A teacher model is used in~\cite{romero2014fitnets,shenteacher} to train a compressed student network. Most similar to our work is \cite{hinton2015distilling}, where they introduce distillation loss for training a model from an ensemble of its own. We show that Label Refinery must be done at the crop level, it benefits from being performed iteratively, and models benefit by learning off of the labels generated by the exact same model.
\vspace{-3mm}
\section{Label Refinery} 
\vspace{-3mm}
\par
Previous works have shown that data augmentation using cropping significantly improves the performance of classification models~\cite{krizhevsky2012imagenet,szegedy2015going}. Given a dataset $\mathcal{D}=\{(X_i, Y_i)\}$, we can formalize data augmentation by defining a new dataset $\tilde{\mathcal{D}} = \{(f({X_i}), Y_i)\}$, where $f$ is a stochastic function that generates crops on-the-fly for the image $X_i$. The image labels assigned to the augmented crops are often not accurate (Figure~\ref{fig:bad-crop-labels} and Figure~\ref{fig:crop-smoothing}). We address this problem by passing the dataset through multiple Label Refiners. The first Label Refinery network $C_{\theta_1}$ is trained over the dataset $\tilde{\mathcal{D}}$ with the inaccurate crop labels. The second Label Refinery network $C_{\theta_2}$ is trained over the same set of images, but uses labels generated by $C_{\theta_1}$. More formally, we can view this procedure as training $C_{\theta_2}$ on a new augmented dataset $\tilde{\mathcal{D}}_1 = \{(f(X_i), C_{\theta_1}(f(X_i)))\}$. Once $C_{\theta_2}$ is trained, we can similarly use it to train a subsequent network $C_{\theta_3}$.


\par
\newcommand{\ptt}[1]{p^{#1}_c(f(X_i))}
\newcommand{\pt}{\ptt{t}}
\newcommand{\ptminusone}{\ptt{t-1}}
We train the first Label Refinery network $C_{\theta_1}$ using the cross-entropy loss against the image-level ground-truth labels. We train all subsequent Label Refinery models $C_{\theta_t}$ for $t>1$ by minimizing the KL-divergence between its output and the soft label generated by the previous Label Refinery $C_{\theta_{t-1}}$. Letting $p^t_c(z) \triangleq C_{\theta_t}(z)[c]$ be the probability assigned to class $c$ in the output of model $C_{\theta_{t}}$ on some crop $z$, our loss function for training model $C_{\theta_t}$ is:
\begin{align}
\label{eq:kl-loss}
    L_t(f(X_i)) &= -\sum _c \ptminusone \log \left( \dfrac{\pt}{\ptminusone} \right) \\
        &= -\sum _c \ptminusone \log \pt + \sum _c \ptminusone \log \ptminusone        \nonumber
\end{align}
The second term is the entropy of the soft labels, and is constant with respect to $C_{\theta_t}$. We can remove it and instead minimize the cross entropy loss:
\begin{align}
    \tilde{L}_t(f(X_i)) &= -\sum _c \ptminusone \log \pt
\end{align}
Note that training $C_{\theta_1}$ using cross entropy loss can be viewed as a special case of our sequential training method using KL-divergence in which $C_{\theta_1}$ is trained from the original image-level labels. It's worth emphasizing that the subsequent models do not see the original ground truth labels $Y_i$. The information in the original labels is propagated by the sequence of Label Refinery networks.

\par
If any of the Label Refinery networks have Batch Normalization~\cite{ioffe2015batch}, we put them in training mode even at the label generation step. That is, their effective mean and standard deviation to be computed from the current training batch as opposed to the saved running mean and running variance. We have observed that this results in more accurate labels and, therefore, more accurate models. We believe that this is due to the fact that the Label Refinery has been trained with the Batch Normalization layers in the training mode. Hence it produces more accurate labels \textit{for the training set} if it's in the same mode.



\par
It is possible to use the same network architecture for some (or all) of the Label Refinery networks in the sequence. We have empirically observed that the dataset labels improve iteratively even when the same network architecture is used multiple times (Section~\ref{sec:experiments}). This is because the same Label Refinery network trained on the new refined dataset becomes more accurate that its previous versions over each pass. Thus, subsequent networks are trained with more accurate labels.

\par
The accuracy of a trained model heavily depends on the consistency of the labels provided to it during training. Unfortunately, assessing the quality of crop labels quantitatively is not possible because there crop level labels are not provided. Asking human annotators to evaluate individual crops is infeasible both due to the number of possible crops and due to the difficulty of evaluating soft labels to a large number of categories for a crop in which there may not be a single main object. We can use a network's validation set accuracy as a measure of its ability to produce correct labels for crops. Intuitively, this measurement serves as an indication of the quality of a Label Refinery network. However, we observe that models with higher validation accuracy do not always produce better crop labels if the model with higher validation accuracy is severely overfit to the training set. Intuitively, this is because the model will reproduce the ground-truth image labels for training set images. We explore this more in Section~\ref{sec:analysis}.


\par
One popular way to augment ImageNet data is to crop patches as small as $8\%$ of the area of the image~\cite{szegedy2015going}. In the presence of such aggressive data augmentation, the original image label is often very inaccurate for the given crop. Whereas traditional methods only augment the image input data through cropping, we additionally augment the labels using Label Refinery networks to produce labels for the crops. Smaller networks such as MobileNet~\cite{howard2017mobilenets} usually aren't trained with such small crops. Yet, we observe that such networks can benefit from small crops if a Label Refinery is used. This demonstrates that a primary cause in accuracy degradation of such networks is inaccurate labels on small crops.

\subsection{Adversarial Jittering}
\label{sec:adversarial}
\par
Using a Label Refinery network allows us to generate labels for any set of images. Our training dataset $\tilde{\mathcal{D}}_t = \{(f(X_i), C_{\theta_t}(f(X_i)))\}$ depends only on the input images $X_i$, and labels are generated on-the-fly by the Refinery network $C_{\theta_t}$. This means that we are no longer limited to using images in the training set $\mathcal{D}$. We could use another unlabeled image dataset as a source of $X_i$. We could even use synthetic images. We experiment with using the Label Refinery in conjunction with the network being trained in order to generate adversarial examples on which the two networks disagree.

\par
Let $C_{\theta_{t-1}}$ and $C_{\theta_t}$ be two of the networks in a sequence of Label Refinery networks. Given a crop $f(X_i)$, we define $\alpha_t(f(X_i))$ to be a modification of $f(X_i)$ for which $C_{\theta_{t-1}}$ and $C_{\theta_t}$ output different probability distributions. Following the practice of \cite{nguyen2015deep} for generating adversarial examples, we define $\alpha_t$ as
\begin{equation}
\label{eq:adversarial}
    \alpha_t(X) = X + \eta \frac{\partial L_t}{\partial X}
\end{equation}
, where $L_t$ is the KL-divergence loss defined in Equation~\ref{eq:kl-loss}. This update performs one step of gradient ascent in the direction of increasing the KL-divergance loss. In other words, the input is modified to exacerbate the discrepancy between the output probability distributions. In order to prevent the model being trained from becoming confused by the unnatural inputs $\alpha_t(f(X_i))$, we batch the adversarial examples with their corresponding natural crops $f(X_i)$.

\newcommand{\explain}{AlexNet is trained off of the ground-truth labels, and the successive models AlexNet\up{$i+1$} are trained off of the labels generated by AlexNet\up{$i$}. }

\section{Experiments}
\label{sec:experiments}
\par
We evaluate the effect of label refining for a variety of network architectures on the standard ImageNet, ILSRVC2012~\cite{deng2009imagenet} classification challenge. We first explore the effect of label refining when the Label Refinery network architecture is identical to the architecture of the network being trained. We then evaluate the effect of label refining when the Label Refinery uses a more accurate network architecture. Finally, we present some ablation studies and analysis to investigate the source of the improvements. Note that all experiments are done with a single model over a single validation crop.

\par
\textbf{Implementation Details:}
All models are trained using PyTorch~\cite{pytorch} on $4$ GPUs for $200$ epochs to ensure convergence. The learning rate is constant for the first $140$ epochs. It is divided by $10$ after epoch $140$ and again divided by $10$ after epoch $170$. We use an initial learning rate of $0.01$ to train AlexNet and an initial learning rate of $0.1$ for all other networks. We use image cropping and horizontal flipping to augment the training set. When cropping, we follow the data augmentation practice of \cite{szegedy2015going} in which the crop areas are chosen uniformly from $8\%$ to $100\%$ of the area of the image. We use a batch size of $256$ for all models except the MobileNet variations, for which we use batch size of $512$. Except for adversarial inputs experiments, we train models from refined labels starting from a random initialization. Our source code is available at {\small\url{http://github.com/hessamb/label-refinery}}.

\begin{table}[t]
\centering
\small
\subfigure{
    \begin{tabular}{|l | c | c |}
        \hline
        \textbf{Model} & \textbf{Top-1} & \textbf{Top-5}\\
        \Xhline{4\arrayrulewidth}
        AlexNet            &  $57.93$      & $79.41$\\
        \hline
        AlexNet\up{2}     &  $59.97$      & $81.44$\\
        \hline
        AlexNet\up{3}     &  $60.87$      & $82.13$\\
        \hline
        AlexNet\up{4}     &  $61.22$      & $\mathbf{82.56}$\\
        \hline
        AlexNet\up{5}     &  $\mathbf{61.37}$      & $\mathbf{82.56}$\\
        \hline
    \end{tabular}
  \label{tab:alexnet-cascade}
}
\subfigure{
    \begin{tabular}{|l | c | c |}
        \hline
        \textbf{Model} & \textbf{Top-1} & \textbf{Top-5}\\
        \Xhline{4\arrayrulewidth}
        ResNet50          &  $75.7$       & $92.81$\\
        \hline
        ResNet50\up{2}    &  $\mathbf{76.5}$      & $\mathbf{93.12}$\\
        \hline
    \end{tabular}
  \label{tab:resnet50-cascade}
}
\subfigure{
    \begin{tabular}{|l | c | c |}
        \hline
        \textbf{Model} & \textbf{Top-1} & \textbf{Top-5}\\
        \Xhline{4\arrayrulewidth}
        MobileNet          &  $68.51$       & $88.13$\\
        \hline
        MobileNet\up{2}    &  $\mathbf{69.52}$       & $\mathbf{88.7}$\\
        \hline
    \end{tabular}
  \label{tab:mobilenet-cascade}
}
\subfigure{
    \begin{tabular}{|l | c | c |}
        \hline
        \textbf{Model} & \textbf{Top-1} & \textbf{Top-5}\\
        \Xhline{4\arrayrulewidth}
        VGG16          &  $70.1$       & $88.54$\\
        \hline
        VGG16\up{2}          &  $71.85$       & $90.07$\\
        \hline
        VGG16\up{3}          &  $\mathbf{72.49}$       & $\mathbf{90.76}$\\
        \hline
    \end{tabular}
  \label{tab:vgg16-cascade}
}
\subfigure{
    \begin{tabular}{|l | c | c |}
        \hline
        \textbf{Model} & \textbf{Top-1} & \textbf{Top-5}\\
        \Xhline{4\arrayrulewidth}
        VGG19          &  $71.39$       & $89.44$\\
        \hline
        VGG19\up{2}          &  $72.66$       & $90.75$\\
        \hline
        VGG19\up{3}          &  $\mathbf{73.32}$       & $\mathbf{91.30}$\\
        \hline
    \end{tabular}
  \label{tab:vgg19-cascade}
}
\subfigure{
    \begin{tabular}{|l | c | c |}
        \hline
        \textbf{Model} & \textbf{Top-1} & \textbf{Top-5}\\
        \Xhline{4\arrayrulewidth}
        Darknet19          &  $70.6$       & $89.13$\\
        \hline
        Darknet19\up{2}          &  $72.74$       & $90.73$\\
        \hline
        Darknet19\up{3}          &  $\mathbf{73.01}$       & $\mathbf{90.92}$\\
        \hline
    \end{tabular}
  \label{tab:darknet19-cascade}
}
\caption{\small Self-Refining results on the ImageNet 2012 validation set. Each model is trained using labels refined by the model right above it. That is, AlexNet\up{3} is trained by the labels refined by AlexNet\up{2}, and AlexNet\up{2} is trained by the labels refined by AlexNet. The first row models are trained using the image level ground-truth labels.}
\label{tab:cascade}
\vspace{-0.7cm}
\end{table}

\par
\textbf{Self-Refinement:}
We first explore using a Label Refinery to train another network with the same architecture. Table~\ref{tab:cascade} shows the results for self-refinement on various architectures. Each row represents a randomly-initialized instance of the network architecture trained with labels refined by the model directly one row above it in the table. All six network architectures improve their accuracy through self-refinement. For AlexNet the self-refining process must be repeated $4$ times before convergence, whereas MobileNet and ResNet-50 converge much faster. We argue that this is because AlexNet is more overfit to the training set. Therefore, it takes more training iterations to forget the information that it has memorized from training examples. One might argue that this is due to the extended training time of models. However, we experimented with training models for an equal number of total epochs and the model accuracies did not improve further. This is discussed further in Section~\ref{sec:analysis}.

\par
\textbf{Cross-Architecture Refinement:}
The architecture of a Label Refinery network can be different from that of the trained network. A high-quality Label Refinery should not overfit on training data even if its validation accuracy is high. In other words, under the same validation accuracy, a network with lower training accuracy is a better Label Refinery.
Intuitively, this property allows the refinery to generate high-quality crop labels that are reflective of the true content of the crops. This property prevents the refinery from simply predicting the training labels. We observe that a ResNet-50 model trained to $75.7\%$ top-1 validation accuracy on ImageNet can serve as a high-quality refinery. Table~\ref{tab:resnet50-refinery} shows that a variety of network architectures benefit significantly from training with refined labels. All network architectures that we tried using Label Refineries gained significant accuracy improvement over their previous state-of-the-art. AlexNet and ResNetXnor-50\footnote{ResNetXnor-50 is the XNOR-net~\cite{rastegari2016xnor} version of ResNet-50 in which layers are binary.} achieve more than a $7$ point improvement in top-1 accuracy. Efficient and compact models such as MobileNet benefit significantly from cross-architecture refinement. VGG networks have a very high capacity and they overfit to the training set more than the other networks. Providing more accurate training set labels helps them to fit to more accurate signals and perform better at validation time. Darknet19, the backbone architecture of YOLOv2~\cite{redmon2017yolo9000}, improves almost $4$ points when trained with refined labels.
\begin{table}[t]
\centering
\scriptsize
\newcolumntype{C}{>{\centering\arraybackslash}p{1.4cm}}
\begin{tabular}{|l | C | C | C | C | C | C |}
    \hline
    \multirow{2}{*}{\textbf{Model}} &
        \multicolumn{2}{|c|}{\textbf{Paper Number}} &
        \multicolumn{2}{|c|}{\textbf{Our Impl.}} &
        \multicolumn{2}{|c|}{\textbf{Label Refinery}} \\
    \cline{2-7}
                                    & Top-1         & Top-5         & Top-1     & Top-5     & Top-1                      & Top-5\\
    \Xhline{4\arrayrulewidth}
    AlexNet~\cite{krizhevsky2012imagenet}          & $59.3$        & $81.8$        & $57.93$   &  $79.41$  & $\mathbf{66.28}^\dagger$   & $\mathbf{86.13}^\dagger$\\
    \hline
    MobileNet~\cite{howard2017mobilenets}      & $70.6$        & \texttt{N/A}  & $68.53$   & $88.14$   & $\mathbf{73.39}$           & $\mathbf{91.07}$\\
    \hline
    MobileNet0.75~\cite{howard2017mobilenets}  & $68.4$        & \texttt{N/A}  & $65.93$   & $86.28$   & $\mathbf{70.92}$           & $\mathbf{89.68}$\\
    \hline
    MobileNet0.5~\cite{howard2017mobilenets}   & $63.7$        & \texttt{N/A}  & $63.03$   & $84.55$   & $\mathbf{66.66}^\dagger$   & $\mathbf{87.07}^\dagger$\\
    \hline
    MobileNet0.25~\cite{howard2017mobilenets}  & $50.6$        & \texttt{N/A}  & $50.65$   & $74.42$   & $\mathbf{54.62}^\dagger$   & $\mathbf{77.92}^\dagger$\\
    \hline
    ResNet-50~\cite{he2016deep}         & \texttt{N/A}  & \texttt{N/A}  & $75.7$    & $92.81$   & $\mathbf{76.5}$            & $\mathbf{93.12}$\\
    \hline
    ResNet-34~\cite{he2016deep}         & \texttt{N/A}  & \texttt{N/A}  & $73.39$   & $91.32$   & $\mathbf{75.06}$           & $\mathbf{92.35}$\\
    \hline
    ResNet-18~\cite{he2016deep}         & \texttt{N/A}  & \texttt{N/A}  & $69.7$    & $89.26$   & $\mathbf{72.52}$           & $\mathbf{90.73}$\\
    \hline
    ResNetXnor-50~\cite{rastegari2016xnor}    & \texttt{N/A}  & \texttt{N/A}  & $63.1$    & $83.61$   & $\mathbf{70.34}$           & $\mathbf{89.18}$\\
    \hline
    VGG16~\cite{simonyan2014very}                & $73$          & $91.2$        & $70.1$    &  $88.54$  & $\mathbf{75}$              & $\mathbf{92.22}$\\
    \hline       
    VGG19~\cite{simonyan2014very}                & $72.7$        & $91$          & $71.39$   &  $89.44$  & $\mathbf{75.46}$           & $\mathbf{92.52}$\\
    \hline
    Darknet19~\cite{redmon2017yolo9000}          & $72.9$        & $91.2$        & $70.6$    &  $89.13$  & $\mathbf{74.47}$           & $\mathbf{91.94}$\\
    \hline
\end{tabular}

\vspace{3mm}
\caption{\small Using refined labels improves the accuracy of a variety of network architectures to new state-of-the-art accuracies. The Label Refinery used in these experiments is a ResNet-50 model trained with weight decay.\\
$^\dagger$ These models can be further improved by training with adversarial inputs (Table~\ref{tab:adversarial-refinery}).}
\label{tab:resnet50-refinery}
\vspace{-0.4cm}
\end{table}

\begin{table}[b!]
\vspace{-5mm}
\centering
\small
\newcolumntype{C}{>{\centering\arraybackslash}p{1.4cm}}
\begin{tabular}{|l | C | C | C | C | C | C |}
    \hline
    \multirow{2}{*}{\textbf{Model}} &
        \multicolumn{2}{|c|}{\textbf{GT Labels}} &
        \multicolumn{2}{|c|}{\textbf{Label Refinery}} &
        \multicolumn{2}{|c|}{\textbf{Adversarial}} \\
    \cline{2-7}
                    & Top-1         & Top-5         & Top-1     & Top-5     & Top-1     & Top-5\\
    \Xhline{4\arrayrulewidth}
    AlexNet         & $57.93$       &  $79.41$      & $66.28$   & $86.13$   & $\mathbf{67.2}$   & $\mathbf{86.92}$\\
    \hline
    MobileNet0.5    & $63.03$       &  $84.55$      & $66.66$   & $87.07$   & $\mathbf{67.33}$  & $\mathbf{87.4}$\\
    \hline
    MobileNet0.25   & $50.65$       &  $74.42$      & $54.62$   & $77.92$   & $\mathbf{55.59}$  & $\mathbf{78.58}$\\
    \hline
\end{tabular}

\vspace{3mm}
\caption{\small Smaller models are further improved by training over adversarial inputs. The Adversarial Label Refinery is ResNet-50.}
\label{tab:adversarial-refinery}
\end{table}

\par
\textbf{Adversarial Inputs:}
As discussed in Section~\ref{sec:adversarial} we can adversarially augment our training set with patches on which the refinery network and the trained model disagree. We used a gradient step of $\eta=1$, as defined in Equation \ref{eq:adversarial} to augment the dataset. We batch each adversarially modified crop with the original crop during training. This helps to ensure the trained model does not drift too far from natural images. We observe in Table~\ref{tab:adversarial-refinery} that smaller models further improve beyond the improvements from using a Label Refinery alone.

\begin{table}[t]
\centering
\small
\newcolumntype{C}{>{\centering\arraybackslash}p{1.4cm}}
\begin{tabular}{|l | C | C | C | C | C | C |}
    \hline
    \textbf{Model}  & Top-1             & Top-5\\
    \Xhline{4\arrayrulewidth}
    AlexNet -- no refinery              & $57.93$        & $79.41$\\
    \hline
    AlexNet -- soft static refinery     & $63.55$        & $84.16$\\
    \hline
    AlexNet -- hard dynamic refinery    & $64.41$        & $84.53$\\
    \hline
    AlexNet -- soft dynamic refinery    & $66.28$        & $86.13$\\
    \hline
\end{tabular}

\vspace{3mm}
\caption{\small AlexNet benefits from both soft labeling and dynamic labeling. When combined the improvement is increased over both, suggesting that they capture different aspects of label errors. Label Refinery is ResNet-50.}
\label{tab:dynamic-vs-soft}
\vspace{-0.4cm}
\end{table}

\begin{table}[b!]
\centering
\small
\newcolumntype{C}{>{\centering\arraybackslash}p{1.4cm}}
\begin{tabular}{|l | C | C | C | C | C | C |}
    \hline
    \textbf{Model}                                      & Top-1             & Top-5\\
    \Xhline{4\arrayrulewidth}
    AlexNet -- no refinery                              & $57.93$           & $79.41$\\
    \hline
    AlexNet -- taxonomy based refined categories        & $56.73$           & $77.69$\\
    \hline
    AlexNet -- visually refined categories              & $58.54$           & $80.77$\\
    \hline
    AlexNet -- visually refined images                  & $62.69$           & $83.46$\\
    \hline
\end{tabular}

\vspace{3mm}
\caption{\small Comparing refining labels at category level vs.\ image level. Note that ``AlexNet -- visually refined images" is trained over image level refined labels as opposed to crop level. For fairness, we fixed the batch normalization layers of label refinery (which harms the quality of label refinery) in all visually refined labels experiments. Label Refinery is ResNet-50.}
\label{tab:category-vs-image}
\end{table}

\subsection{Analysis} 
\label{sec:analysis}
We explore the characteristics of models trained using a Label Refinery. We first explore how much of the improvement comes from the dynamic labeling of the image crops and how much of it comes from softening the target labels. We then explore the overfitting characteristics of models trained with a Label Refinery. Finally, we explore using various loss functions to train models against the refined labels. Most of the analyses are performed on AlexNet architecture because it trains relatively fast ($\sim 1$ day) on the ImageNet dataset.

\par
\textbf{Dynamic Labels vs.\ Soft Labels:}
The benefits of using a label refinery are twofold: (1) Each crop is dynamically re-labeled with more accurate labels for the crop (Figure~\ref{fig:bad-crop-labels}), and (2) images are softly labeled according to the distribution of visually similar objects in the crop (Figure ~\ref{fig:crop-smoothing}). We find that both aspects of the refinement process improve performance. To assess the improvement from dynamic labeling alone, we perform label refinement with hard dynamic labels. Specifically, we assign a one-hot label to each crop by passing the crop to the Label Refinery and choosing the most-likely category from the output. To observe the improvement from soft labeling alone, we perform label refinement with soft static labels. To compute these labels for a given crop, we pass a center crop of the original image to the refiner rather than using the training crop. We compare the results for soft static labels and hard dynamic labels in Table~\ref{tab:dynamic-vs-soft}. Both dynamic labeling and soft labeling significantly improve the accuracy of AlexNet. When they are combined we observe an additional improvement, suggesting that they address different issues with labels in the dataset.

\begin{figure}[t]
  \centering
  \subfigure[]{
    \includegraphics[width=0.45\textwidth]{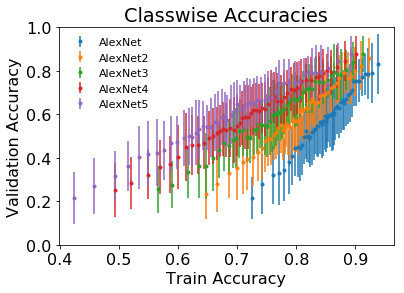}
    \label{fig:overfitting_alex}
  }
  \subfigure[]{
    \includegraphics[width=0.45\textwidth]{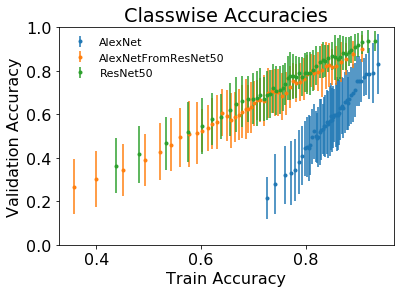}
    \label{fig:overfitting_resnet}
  }
  \label{fig:overfitting_ar}
  \caption{
  Per category train and test accuracy. For each model, labels were sorted according to training set accuracies and divided into bins. each point in the plot shows the average validation set accuracy and the associated standard deviation for each bin. These figures show that training with a refinery results in models with less overfitting.
  }
  \vspace{-0.4cm}
\end{figure}

\begin{figure}[b!]
    \centering
    \includegraphics[width=\textwidth]{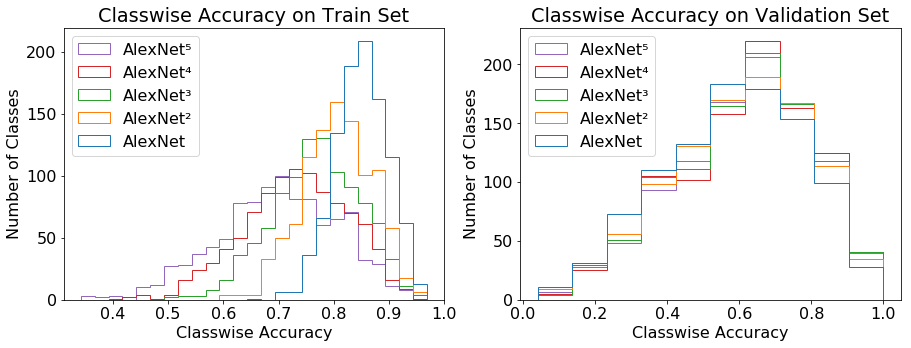}
    \caption{The train and validation accuracy distribution of AlexNet models trained sequentially. \explain}
    \label{fig:trainval-accuracies-alexnet}
\end{figure}

\par
\textbf{Category Level Refining vs.\ Image Level Refining: }
Labels can be refined at the category level. That is, all images in a class can be assigned a unique soft label that models intra-category similarities. At the category level, labels can be refined either by visual cues (based on the visual similarity between the categories) or by semantic relations (based on the taxonomic relationship between the categories). Since ImageNet categories are drawn from WordNet, we can use taxonomy-based distances to refine the labels. We experiment with using the Wu-Palmer similarity~\cite{wu1994verbs} of the WordNet~\cite{miller1990introduction} categories to refine the category labels. Table~\ref{tab:category-vs-image} compares refining labels at the category level with refining at the image level. We observe larger improvements when the labels are refined at the image level. Our experiment shows that taxonomy-based refinement does not improve training. We believe this is because WordNet similarities do not correlate well with visual similarities in the image space. Refining category labels based off of their WordNet distance can confuse the target model.

\begin{figure}[t]
    \centering
    \includegraphics[width=\textwidth]{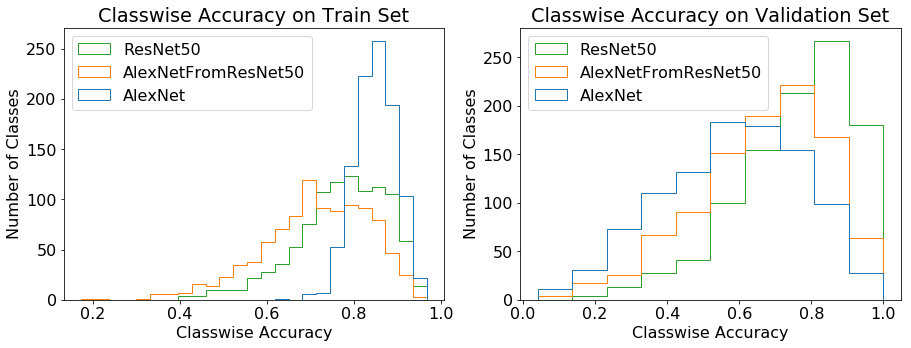}
    \caption{
    The train and validation accuracies for AlexNet, ResNet, and AlexNet trained off of labels generate by ResNet50. \texttt{AlexNetFromResNet50} has a train accuracy profile that more closely resembles ResNet50 than AlexNet.}
    \label{fig:trainval-accuracies-resnet}
    \vspace{-0.4cm}
\end{figure}

\begin{table}[b]
\centering
\small
\newcolumntype{C}{>{\centering\arraybackslash}p{1.4cm}}
\begin{tabular}{|l | C | C | C | C |}
    \hline
    \multirow{2}{*}{\textbf{Model}} &
        \multicolumn{2}{|c|}{\textbf{Refinery}} &
        \multicolumn{2}{|c|}{\textbf{AlexNet}}\\
    \cline{2-5}
                                        & Top-1          & Top-5         & Top-1     & Top-5\\
    \Xhline{4\arrayrulewidth}
    AlexNet -- no refinery              & \texttt{N/A}   & \texttt{N/A}  & $57.93$        & $79.41$\\
    \hline
    AlexNet -- refinery: VGG16          & $70.1$         & $88.54$       & $60.78$        & $81.80$\\
    \hline
    AlexNet -- refinery: MobileNet      & $68.53$        & $88.14$       & $65.22$        & $85.69$\\
    \hline
    AlexNet -- refinery: ResNet-50      & $75.7$         & $92.81$       & $66.28$        & $86.13$\\
    \hline
\end{tabular}

\vspace{3mm}
\caption{\small Different architecture choices for the refinery network.}
\label{tab:arch-choices}
\end{table}

\par
\textbf{Model Generalization:}
Figure~\ref{fig:overfitting_alex}~and~\ref{fig:overfitting_resnet} show the per-category train and validation accuracies of ImageNet categories for models trained with a Label Refinery. Each point in the plot shows the average and standard deviation of the accuracies for a set of categories.  Figure~\ref{fig:overfitting_alex} shows the accuracies of a sequence of AlexNet models (in different colors). AlexNet trained using the ground-truth labels has much higher train accuracy. Successive models demonstrate less overfitting as shown by the decrease in the ratio between train accuracy and validation accuracy. Figure~\ref{fig:overfitting_resnet} shows the per-category accuracies of AlexNet and ResNet-50, as well as an AlexNet model trained with a ResNet50 Label Refinery. ResNet-50 trained with weight decay generalizes better compared to AlexNet, which has two fully connected layers. Intuitively, the generalization of ResNet-50 enbles it to generate accurate per-crop labels for the training set. Thus, training AlexNet with a ResNet-50 Label Refinery  allows AlexNet to perform well on the test set without overfiting to the original ground-truth labels.

\begin{table}[t]
\centering
\small
\newcolumntype{C}{>{\centering\arraybackslash}p{1.4cm}}
\begin{tabular}{|l | C | C |}
    \hline
    \textbf{Model}  & Top-1             & Top-5\\
    \Xhline{4\arrayrulewidth}
    AlexNet -- no refinery              & $57.93$        & $79.41$\\
    \hline
    AlexNet -- l\sub{2} loss            & $63.16$        & $85.56$\\
    \hline
    AlexNet -- KL-divergence from output to label    & $65.36$        & $85.41$\\
    \hline
    AlexNet -- KL-divergence from label to output    & $66.28$        & $86.13$\\
    \hline
\end{tabular}

\vspace{3mm}
\caption{\small Different loss function choices. Label Refinery is ResNet-50.}
\label{tab:loss-choices}
\vspace{-0.4cm}
\end{table}

\begin{figure}[b!] 
    \centering
    \includegraphics[height=.45\textwidth]{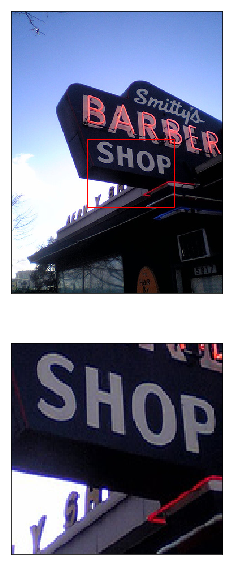}
    \includegraphics[height=.4\textwidth]{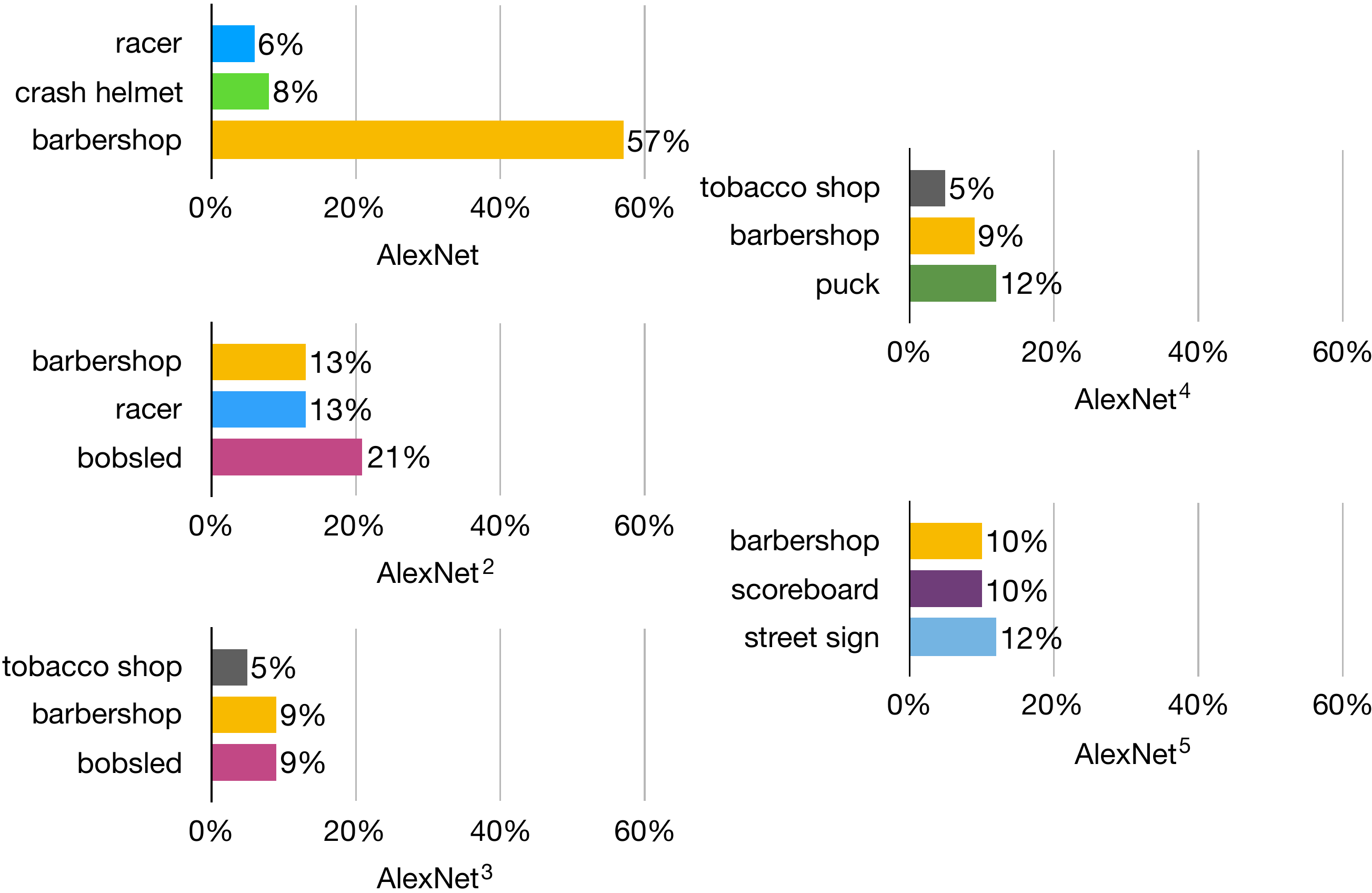}
    \caption{\footnotesize The top three predictions for a crop of an image labelled ``barber shop'' in the ImageNet training set. AlexNet trained on the ground-truth labels is overfit towards the image level label. Successive AlexNet models overfit less, reducing the weight of the ``barber shop'' category and eventually assigning more probability to other plausible categories such as ``street sign'' and ``scoreboard''.}
    \label{fig:barbershop}
\end{figure}

\par
Figure~\ref{fig:trainval-accuracies-alexnet} shows the training set and validation set accuracies of a sequence of AlexNet models trained with a Label Refinery. The AlexNet trained with ground-truth labels achieves $\sim 86\%$ training accuracy for the majority of classes, but achieves much lower validation set accuracies. By contrast, AlexNet\up{5} has a training accuracy profile more closely resembling its validation accuracy profile. Figure~\ref{fig:trainval-accuracies-resnet} shows a similar phenomena training AlexNet with a ResNet-50 refinery. It's interesting to note that the training and validation profiles of AlexNet trained with a ResNet50 Label Refinery more closely resemble the refinery than the original AlexNet.

\par
\textbf{Choice of Label Refinery Network:}
A good Label Refinery network should generate accurate labels for the \textit{training set} crops. A Label Refinery's validation accuracy is an informative signal of its quality. However, if the Label Refinery network is heavily overfitted on the training set, it will not be helpful during training because it will produce the same ground-truth label for all image crops. Table~\ref{tab:arch-choices} compares different architecture choices for refinery network. VGG16 is a worse choice of Label Refinery than MobileNet, even though VGG16 is more accurate. This is because VGG16 severely overfits to the training set and therefore produces labels too similar to the ground-truth.

\par
\textbf{Choice of Loss Function:}
We can use a variety of loss functions to train our target networks to match the soft labels. The KL-divergence loss function that we use is a generalization of the standard cross-entropy classification loss. Note that KL-divergence is not a symmetric function (\ie $D_{KL}(P||Q) \neq D_{KL}(Q||P)$). Table~\ref{tab:loss-choices} shows the model accuracy if other standard loss functions are used.

\par
\begin{figure}[t!]
    \centering
    \includegraphics[height=.2\textwidth]{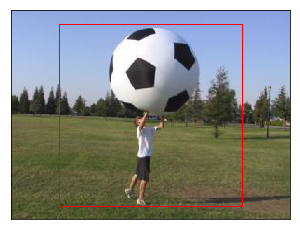}
    \includegraphics[height=.22\textwidth]{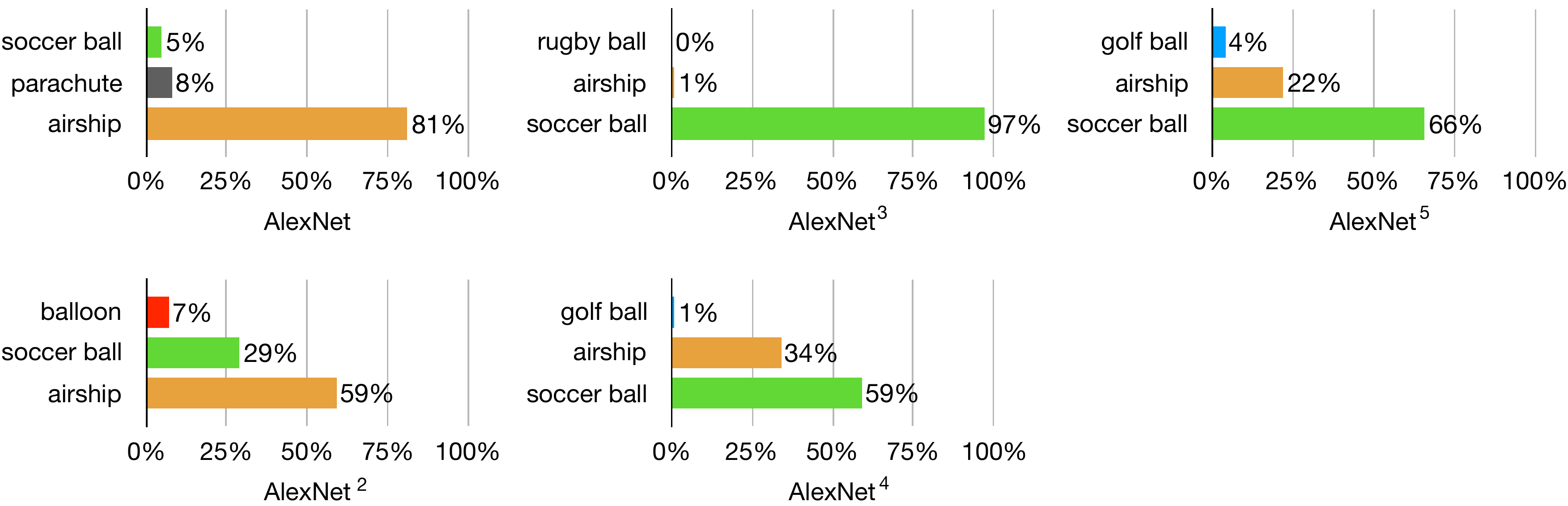}
    \caption{The top three predictions for an image labelled ``soccer ball'' in the ImageNet validation set. Successive models learn to avoid overfitting an object surrounded by patches of sky to the ``airship'' category.}
    \label{fig:soccer-ball}
    \vspace{-4mm}
\end{figure}

\textbf{Qualitative Results:}
Using a refinery to produce crop labels reduces overfitting by providing more accurate labels during training. In Figure~\ref{fig:barbershop}, we see an example in which a training image crop does not contain enough information to identify the image category as ``barbershop''. In spite of this, AlexNet assigns the crop a label of barbershop with high confidence. This is due to overfitting on the training set. By using an AlexNet as a refinery, AlexNet\up{2} learns to generalize better. It produces a lower score for ``barbershop'', and a higher score for other categories. Generalization behavior improves with successive rounds of label refining until AlexNet\up{5} produces a smooth distribution over plausible categories.
In Figure~\ref{fig:soccer-ball}, we see an example of a ``soccer ball'' from the validation set of ImageNet. AlexNet incorrectly predicts ``airship'' with high confidence. This prediction is most likely because the main object is surrounded by blue sky, which is common for an airship but uncommon for a soccer ball. By using AlexNet as a refinery to train another AlexNet model we achieve a reduced score for ``airship'' and a higher score for ``soccer ball''. After several rounds of successive refining we achieve an AlexNet model that makes the correct prediction without completely forgetting the similarities between the soccer ball in the sky and an airship.

\section{Conclusion}
In this paper we address shortcomings commonly found in the labels of supervised learning pipelines. We introduce a solution to refine the labels during training in order to improve the generalization and the accuracy of learning models. The proposed Label Refinery allows us to dynamically label augmented training crops with soft targets. Using a Label Refinery, we achieve a significant gain in the classification accuracy across a wide range of network architectures. Our experimental evaluation shows improvement in the state-of-the-art accuracy for popular architectures including AlexNet, VGG, ResNet, MobileNet, and XNOR-Net.

\bibliographystyle{splncs}
\bibliography{bib}
\end{document}